\pdfoutput=1

\documentclass[11pt]{article}

\usepackage[]{ACL2023}

\usepackage{times}
\usepackage{latexsym}
\usepackage{booktabs}
\usepackage{multirow}

\usepackage[T1]{fontenc}

\usepackage[utf8]{inputenc}

\usepackage{microtype}
\usepackage{graphicx}
\usepackage{amsmath}
\usepackage{subfigure}
\usepackage{ulem}

\usepackage{inconsolata}

\title{Towards Better Entity Linking with Multi-View Enhanced Distillation}
\author{Yi Liu$^{1,2,}$\footnotemark[1], Yuan Tian$^{3}$, Jianxun Lian$^{3}$, Xinlong Wang$^{3}$, Yanan Cao$^{1,2,}$\footnotemark[2],\\ {\bf Fang Fang$^{1,2}$,}  {\bf Wen Zhang$^{3}$,} {\bf Haizhen Huang$^{3}$,} {\bf Denvy Deng$^{3}$,} {\bf Qi Zhang$^{3}$}\\
  $^{1}$Institute of Information Engineering, Chinese Academy of Sciences \\
  $^{2}$School of Cyber Security, University of Chinese Academy of Sciences \\
  $^{3}$Microsoft \\
  \texttt{\{liuyi1999,caoyanan,fangfang0703\}@iie.ac.cn} \\
  \texttt{\{yuantian,jialia,xinlongwang,zhangw,hhuang,dedeng,qizhang\}@microsoft.com}}

\begin{document}
\maketitle
\renewcommand{\thefootnote}{\fnsymbol{footnote}}
\footnotetext[1]{Work is done during internship at Microsoft.}
\footnotetext[2]{Corresponding Author.}
\renewcommand{\thefootnote}{\arabic{footnote}}
\begin{abstract}
Dense retrieval is widely used for entity linking to retrieve entities from large-scale knowledge bases. Mainstream techniques are based on a dual-encoder framework, which encodes mentions and entities independently and calculates their relevances via rough interaction metrics, resulting in difficulty in explicitly modeling multiple mention-relevant parts within entities to match divergent mentions. Aiming at learning entity representations that can match divergent mentions, this paper proposes a \textbf{M}ulti-\textbf{V}iew Enhanced \textbf{D}istillation (MVD) framework, which can effectively transfer knowledge of multiple fine-grained and mention-relevant parts within entities from cross-encoders to dual-encoders. Each entity is split into multiple views to avoid irrelevant information being over-squashed into the mention-relevant view. We further design cross-alignment and self-alignment mechanisms for this framework to facilitate fine-grained knowledge distillation from the teacher model to the student model. Meanwhile, we reserve a global-view that embeds the entity as a whole to prevent dispersal of uniform information. Experiments show our method achieves state-of-the-art performance on several entity linking benchmarks\footnote{Our code is available at \url{https://github.com/Noen61/MVD}}. 

\end{abstract}

\section{Introduction}
Entity Linking~(EL) serves as a fundamental task in Natural Language Processing~(NLP), connecting mentions within unstructured contexts to their corresponding entities in a Knowledge Base~(KB). EL usually provides the entity-related data foundation for various tasks, such as KBQA \cite{ye-etal-2022-rng}, Knowledge-based Language Models \citep{liu2020k} and Information Retrieval \citep{li2022cooperative}. Most EL systems consist of two stages: entity retrieval~(candidate generation), which retrieves a small set of candidate entities corresponding to mentions from a large-scale KB with low latency, and entity ranking~(entity disambiguation), which ranks those candidates using a more accurate model to select the best match as the target entity. This paper focuses on the entity retrieval task, which poses a significant challenge due to the need to retrieve targets from a large-scale KB. Moreover, the performance of entity retrieval is crucial for EL systems, as any recall errors in the initial stage can have a significant impact on the performance of the latter ranking stage \citep{luan2021sparse}.

\begin{figure}[t]   
\includegraphics[width=8cm]{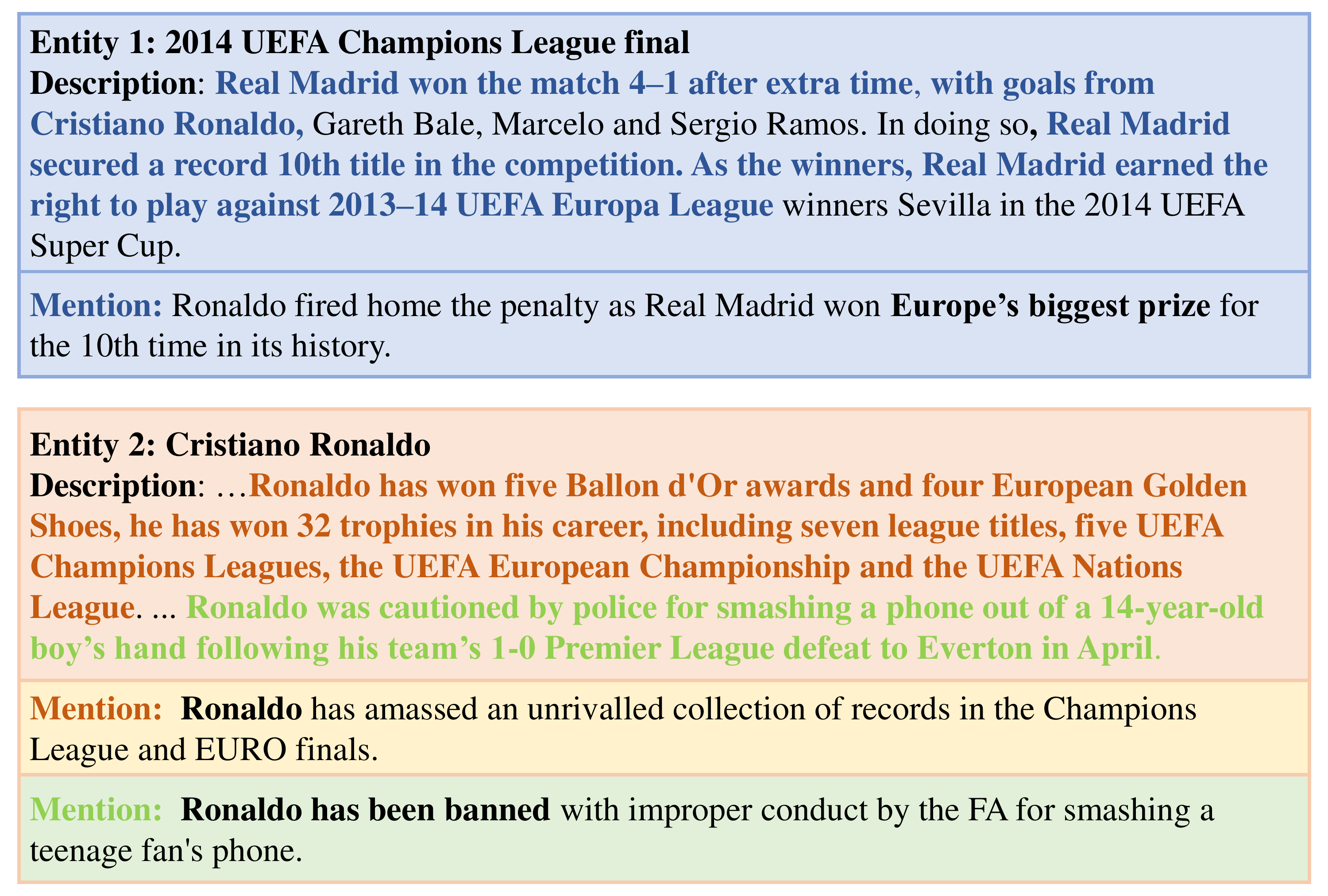}
\caption{The illustration of two types of entities. Mentions in contexts are in \textbf{bold}, key information in entities is highlighted in color. The information in the first type of entity is relatively consistent and can be matched with a corresponding mention. In contrast, the second type of entity contains diverse and sparsely distributed information, can match with divergent mentions.}
\label{fig1}
\end{figure}

Recent advancements in pre-trained language models~(PLMs) \citep{kenton2019bert} have led to the widespread use of dense retrieval technology for large-scale entity retrieval \citep{gillick2019learning,wu2020scalable}. This approach typically adopts a dual-encoder architecture that embeds the textual content of mentions and entities independently into fixed-dimensional vectors \citep{karpukhin2020dense} to calculate their relevance scores using a lightweight interaction metric~(e.g., dot-product). This allows for pre-computing the entity embeddings, enabling entities to be retrieved through various fast nearest neighbor search techniques \citep{johnson2019billion,jayaram2019diskann}. 

The primary challenge in modeling relevance between an entity and its corresponding mentions lies in explicitly capturing the mention-relevant parts within the entity. By analyzing the diversity of intra-information within the textual contents of entities, we identify two distinct types of entities, as illustrated in Figure \ref{fig1}.  Entities with uniform information can be effectively represented by the dual-encoder; however, due to its single-vector representation and coarse-grained interaction metric, this framework may struggle with entities containing divergent and sparsely distributed information. To alleviate the issue, existing methods construct multi-vector entity representations from different perspectives~\citep{ma2021muver,zhang2021understanding,tang2021bidirectional}. Despite these efforts, all these methods rely on coarse-grained entity-level labels for training and lack the necessary supervised signals to select the most relevant representation for a specific mention from multiple entity vectors. As a result, their capability to effectively capture multiple fine-grained aspects of an entity and accurately match mentions with varying contexts is significantly hampered, ultimately leading to suboptimal performance in dense entity retrieval.

In order to obtain fine-grained entity representations capable of matching divergent mentions, we propose a novel Multi-View Enhanced Distillation (MVD) framework. MVD effectively transfers knowledge of multiple fine-grained and mention-relevant parts within entities from cross-encoders to dual-encoders. By jointly encoding the entity and its corresponding mentions, cross-encoders enable the explicit capture of mention-relevant components within the entity, thereby facilitating the learning of fine-grained elements of the entity through more accurate soft-labels. To achieve this, our framework constructs the same multi-view representation for both modules by splitting the textual information of entities into multiple fine-grained views. This approach prevents irrelevant information from being over-squashed into the mention-relevant view, which is selected based on the results of cross-encoders.

We further design cross-alignment and self-alignment mechanisms for our framework to separately align the original entity-level and fine-grained view-level scoring distributions, thereby facilitating fine-grained knowledge transfer from the teacher model to the student model. Motivated by prior works~\citep{xiong2020approximate,zhan2021optimizing,qu2021rocketqa,ren2021rocketqav2}, MVD jointly optimizes both modules and employs an effective hard negative mining technique to facilitate transferring of hard-to-train knowledge in distillation. Meanwhile, we reserve a global-view that embeds the entity as a whole to prevent dispersal of uniform information and better represent the first type of entities in Figure \ref{fig1}.

Through extensive experiments on several entity linking benchmarks, including ZESHEL, AIDA-B, MSNBC, and WNED-CWEB, our method demonstrates superior performance over existing approaches. The results highlight the effectiveness of MVD in capturing fine-grained entity representations and matching divergent mentions, which significantly improves entity retrieval performance and facilitates overall EL performance by retrieving high-quality candidates for the ranking stage.

\section{Related Work}
To accurately and efficiently acquire target entities from large-scale KBs, the majority of EL systems are designed in two stages: entity retrieval and entity ranking. For entity retrieval, prior approaches typically rely on simple methods like frequency information \citep{yamada2016joint}, alias tables \citep{fang2019joint} and sparse-based models \citep{robertson2009probabilistic} to retrieve a small set of candidate entities with low latency. For the ranking stage, neural networks had been widely used for calculating the relevance score between mentions and entities \citep{yamada2016joint,ganea2017deep,fang2019joint,kolitsas2018end}. 

Recently, with the development of PLMs \citep{kenton2019bert,lewis2020bart}, PLM-based models have been widely used for both stages of EL. \citet{logeswaran2019zero} and \citet{yao2020zero} utilize the cross-encoder architecture that jointly encodes mentions and entities to rank candidates,  \citet{gillick2019learning} employs the dual-encoder architecture for separately encoding mentions and entities into high-dimensional vectors for entity retrieval. BLINK \citep{wu2020scalable} improves overall EL performance by incorporating both architectures in its retrieve-then-rank pipeline, making it a strong baseline for the task. GERENE \citep{de2020autoregressive} directly generates entity names through an auto-regressive approach.

To further improve the retrieval performance, various methods have been proposed. \citet{zhang2021understanding} and \citet{sun2022transformational} demonstrate the effectiveness of hard negatives in enhancing retrieval performance. \citet{agarwal2022entity} and GER \citep{wu2023modeling} construct mention/entity centralized graph to learn the fine-grained entity representations. However, being limited to the single vector representation, these methods may struggle with entities that have multiple and sparsely distributed information. Although \citet{tang2021bidirectional} and MuVER \citep{ma2021muver} construct multi-view entity representations and select the optimal view to calculate the relevance score with the mention, they still rely on the same entity-level supervised signal to optimize the scores of different views within the entity, which limit the capacity of matching with divergent mentions. 

In contrast to existing methods, MVD is primarily built upon the knowledge distillation technique~\citep{hinton2015distilling}, aiming to acquire fine-grained entity representations from cross-encoders to handle diverse mentions. To facilitate fine-grained knowledge transfer of multiple mention-relevant parts, MVD splits the entity into multiple views to avoid irrelevant information being squashed into the mention-relevant view, which is selected by the more accurate teacher model. This Framework further incorporates cross-alignment and self-alignment mechanisms to learn mention-relevant view representation from both original entity-level and fine-grained view-level scoring distributions, these distributions are derived from the soft-labels generated by the cross-encoders.

\section{Methodology}

\subsection{Task Formulation}
We first describe the task of entity linking as follows. Give a mention $m$ in a context sentence $s = <c_l,m,c_r>$, where $c_l$ and $c_r$ are words to the left/right of the mention, our goal is to efficiently obtain the entity corresponding to $m$ from a large-scale entity collection $\varepsilon = \lbrace e_1,e_2,...,e_N \rbrace$, each entity $e\in \varepsilon$ is defined by its title $t$ and description $d$ as a generic setting in neural entity linking \citep{ganea2017deep}.  Here we follow the two-stage paradigm proposed by \citep{wu2020scalable}: 1) retrieving a small set of candidate entities $\lbrace e_1,e_2,...,e_K \rbrace $ corresponding to mention $m$ from $\varepsilon$, where $K\ll N$; 2) ranking those candidates to obtain the best match as the target entity. In this work, we mainly focus on the first-stage retrieval.

\subsection{Encoder Architecture}
In this section, we describe the model architectures used for dense retrieval. Dual-encoder is the most adopted architecture for large-scale retrieval as it separately embeds mentions and entities into high-dimensional vectors, enabling offline entity embeddings and efficient nearest neighbor search. In contrast, the cross-encoder architecture performs better by computing deeply-contextualized representations of mention tokens and entity tokens, but is computationally expensive and impractical for first-stage large-scale retrieval \citep{reimers2019sentence,humeau2019poly}. Therefore, in this work, we use the cross-encoder only during training, as the teacher model, to enhance the performance of the dual-encoder through the distillation of relevance scores.

\subsubsection{Dual-Encoder Architecture} 
Similar to the work of \citep{wu2020scalable} for entity retrieval, the retriever contains two-tower PLM-based encoders $\rm Enc_m(\cdot)$ and $\rm Enc_e(\cdot)$ that encode mention and entity into single fixed-dimension vectors independently, which can be formulated as:
\begin{equation} \label{eq1}
\begin{aligned}
     & {E(m)} = {\rm Enc_m}({\rm [CLS]}\, \rm c_l \,{\rm [M_s]}\, \rm m \,{\rm [M_e]}\, c_r \,{\rm [SEP]}) \\
     & {E(e)} = {\rm Enc_e}({\rm [CLS]} \, \rm t \, {\rm [ENT]} \, d \, {\rm [SEP]})
\end{aligned}
\end{equation}
where $\rm m$,$\rm c_l$,$\rm c_r$,$\rm t$, and $\rm d$ are the word-piece tokens of the mention, the context before and after the mention, the entity title, and the entity description. The special tokens $\rm [M_s]$ and $\rm [M_e]$ are separators to identify the mention, and $\rm [ENT]$ serves as the delimiter of titles and descriptions. $\rm [CLS]$ and $\rm [SEP]$ are special tokens in BERT.
For simplicity, we directly take the $\rm [CLS]$ embeddings $ E(m)$ and $ E(e)$ as the representations for mention $m$ and entity $e$, then the relevance score $s_{de}(m,e)$ can be calculated by a dot product $s_{de}(m,e) = E(m) \cdot E(e)$.
\subsubsection{Cross-Encoder Architecture}
Cross-encoder is built upon a PLM-based encoder $\rm Enc_{ce}(\cdot)$, which concatenates and jointly encodes mention $m$ and entity $e$ (remove the $\rm [CLS]$ token in the entity tokens), then gets the $\rm [CLS]$ vectors as their relevance representation ${\rm E}(m,e)$, finally fed it into a multi-layer perceptron (MLP) to compute the relevance score $s_{ce}(m,e)$.

\subsubsection{Multi-View Based Architecture} \label{sec3.2.3}
 With the aim to prevent irrelevant information being over-squashed into the entity representation and better represent the second type of entities in Figure \ref{fig1}, we construct multi-view entity representations for the entity-encoder $\rm Enc_e(\cdot)$. The textual information of the entity is split into multiple fine-grained \textbf{local-views} to explicitly capture the key information in the entity and match mentions with divergent contexts. Following the settings of MuVER \citep{ma2021muver},  for each entity $e$, we segment its description $d$ into several sentences $d^t(t=1,2,..,n)$ with NLTK toolkit \footnote{\url{www.nltk.org}}, and then concatenate with its title $t$ as the $t$-th view $e^t(t=1,2,..,n)$:
\begin{equation} \label{eq2}
\begin{aligned}
      {E(e^t)} = {\rm Enc_e}({\rm [CLS]} \, \rm t \, {\rm [ENT]} \, d^t \, {\rm [SEP]}) 
\end{aligned}
\end{equation}
Meanwhile, we retain the original entity representation $E(e)$ defined in Eq. \eqref{eq1}  as the \textbf{global-view} $\boldsymbol{e^0}$ \textbf{in inference}, to avoid the uniform information being dispersed into different views and better represent the first type of entities in Figure \ref{fig1}. Finally, the relevance score $s(m,e_i)$ of mention $m$ and entity $e_i$ can be calculated with their multiple embeddings. Here we adopt a max-pooler to select the view with the highest relevant score as the \textbf{mention-relevant view}:

\begin{equation}
\begin{aligned}
    s(m,e_i) &=  \underset{t}{\max}\lbrace s(m,e^t_i)\rbrace \\
    &= \underset{t}{\max}\lbrace E(m)\cdot E(e^t)\rbrace
\end{aligned}    
\end{equation} 
\begin{figure*}[t]   
\includegraphics[width=16.5cm]{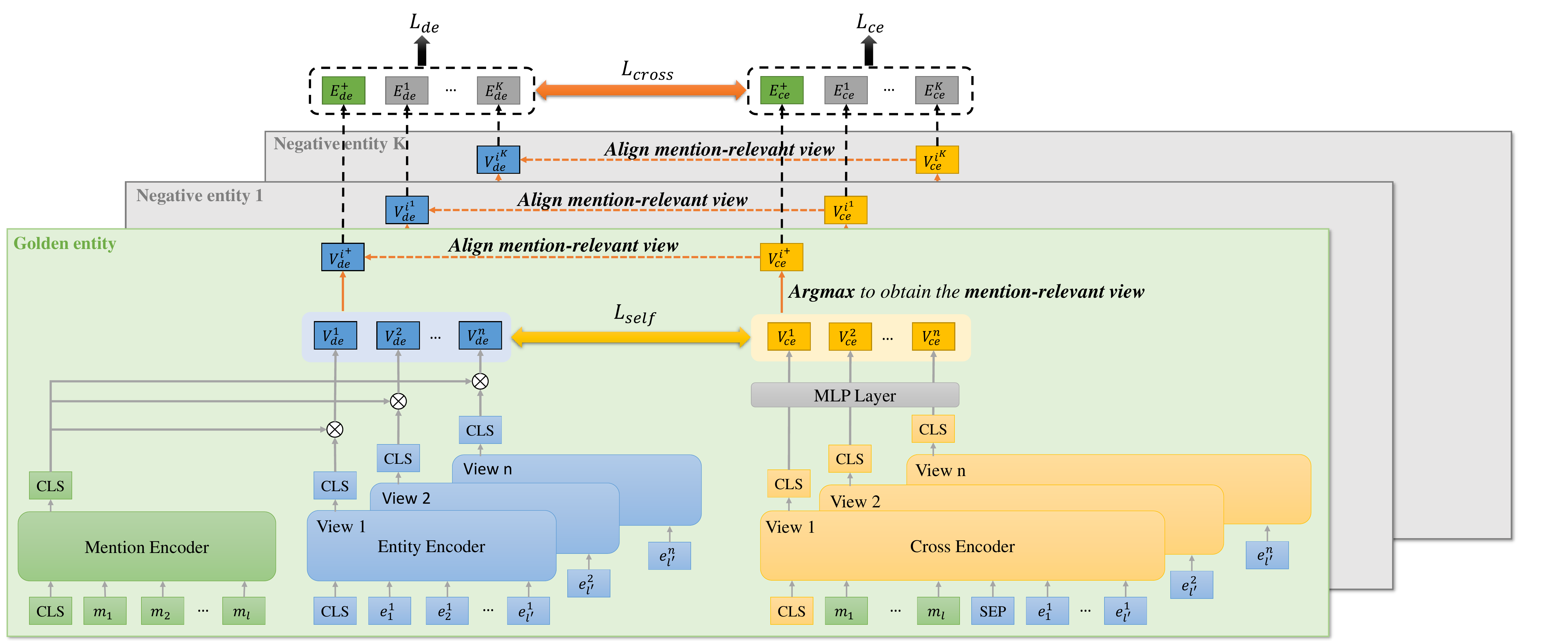}
\caption{The general framework of Multi-View Enhanced Distillation (MVD).  $V^i_{de}$ and $V^i_{ce}$ are the relevance scores between $m$ and $e^i$ separately calculated by dual-encoder and cross-encoder, $E_{de}$ and $E_{ce}$ are the entity relevance scores represented by $V^i_{de}$ and $V^i_{ce}$, base on the max-score view's index \textbf{i} in cross-encoder.}
\label{fig2}
\end{figure*}
\subsection{Multi-View Enhanced Distillation} \label{sec3.3}
The basic intuition of MVD is to accurately transfer knowledge of multiple fine-grained views from a more powerful cross-encoder to the dual-encoder to obtain mention-relevant entity representations.  First, in order to provide more accurate relevance between mention $m$ and each view $e^t(t=1,2,...,n)$ of the entity $e$ as a supervised signal for distillation, we introduce a multi-view based cross-encoder following the formulation in Sec \ref{sec3.2.3}:
\begin{equation}
\begin{aligned}
      {E(m,e^t)} = {\rm Enc_{ce}}({\rm [CLS]} \, \rm m_{enc} \, {\rm [SEP]} \, e^t_{enc} \, {\rm [SEP]}) 
\end{aligned}
\end{equation}
where $\rm m_{enc}$ and $\rm e^t_{enc}(t=1,2,..,n)$ are the word-piece tokens of the mention and entity representations defined as in Eq. \eqref{eq1} and \eqref{eq2}, respectively. 

We further design cross-alignment and self-alignment mechanisms to separately align the original entity-level scoring distribution and fine-grained view-level scoring distribution, in order to facilitate the fine-grained knowledge distillation from the teacher model to the student model.

\noindent \textbf{Cross-alignment} In order to learn entity-level scoring distribution among candidate entities at the multi-view scenario, we calculate the relevance score $s(m,e_i)$ for mention $m$ and candidate entity $e_i$ in candidates $\lbrace e_1,e_2,...,e_K \rbrace $ by all its views $\lbrace e_i^1,e_i^2,...,e_i^n \rbrace $, the indexes of relevant views $i_{de}$ and $i_{ce}$ for dual-encoder and cross-encoder are as follows:
\begin{equation}
\begin{aligned}
    & i_{de} =  \underset{t}{\arg \max}\lbrace s_{de}(m,e^t_i)\rbrace \\
    & i_{ce} =  \underset{t}{\arg \max}\lbrace s_{ce}(m,e^t_i)\rbrace \\
\end{aligned}    
\end{equation}
here to avoid the mismatch of relevant views (i.e., $i_{de} \neq i_{ce}$), we \textbf{align their relevant views} based on the index $i_{ce}$ of max-score view in cross-encoder, the loss can be measured by KL-divergence as
\begin{equation} \label{eq8}
\begin{aligned}
    & \mathcal{L}_{cross} = \sum\limits_{i=1}^{K} {\tilde{s}_{ce}(m,e_i)} \cdot log\frac{{\tilde{s}_{ce}(m,e_i)}}{{\tilde{s}_{de}(m,e_i)}} 
\end{aligned} 
\end{equation}
where
\begin{equation}
\begin{aligned}
    & {\tilde{s}_{de}(m,e_i)}= \frac{e^{s_{de}(m,e_i^{i_{ce}})}}{e^{s_{de}(m,e_i^{i_{ce}})}+\sum\limits_{j\neq i}e^{s_{de}(m,e_j^{j_{ce}})}} \\
    & {\tilde{s}_{ce}(m,e_i)}= \frac{e^{s_{ce}(m,e_i^{i_{ce}})}}{e^{s_{ce}(m,e_i^{i_{ce}})}+\sum\limits_{j\neq i}e^{s_{ce}(m,e_j^{j_{ce}})}} \\
\end{aligned} 
\end{equation}
here $\tilde{s}_{de}(m,e_i)$ and $\tilde{s}_{ce}(m,e_i)$ denote the probability  distributions of the entity-level scores which are represented by the $i_{ce}$-th view over all candidate entities.

\noindent \textbf{Self-alignment} 
Aiming to learn the view-level scoring distribution within each entity for better distinguishing relevant view from other irrelevant views, we calculate the relevance score $s(m,e^t)$ for mention $m$ and each view $e^t_i(t=1,2,...,n)$ of entity $e_i$, the loss can be measured by KL-divergence as:
\begin{equation} \label{eq9}
\begin{aligned}
    & \mathcal{L}_{self} = \sum\limits_{i=1}^{K} \sum\limits_{t=1}^{n} {\tilde{s}_{ce}(m,e_i^t)} \cdot log\frac{{\tilde{s}_{ce}(m,e_i^t)}}{{\tilde{s}_{de}(m,e_i^t)}}
\end{aligned} 
\end{equation}
where
\begin{equation}
\begin{aligned}
    & {\tilde{s}_{de}(m,e_i^t)}= \frac{e^{s_{de}(m,e_i^t)}}{e^{s_{de}(m,e_i^t)}+\sum\limits_{j\neq t}e^{s_{de}(m,e_i^j)}} \\
    & {\tilde{s}_{ce}(m,e_i^t)}= \frac{e^{s_{ce}(m,e_i^t)}}{e^{s_{ce}(m,e_i^t)}+\sum\limits_{j\neq t}e^{s_{ce}(m,e_i^j)}} 
\end{aligned} 
\end{equation}
here $\tilde{s}_{de}(m,e_i^t)$ and $\tilde{s}_{ce}(m,e_i^t)$ denote the probability distributions of the view-level scores over all views within each  entity.

\noindent \textbf{Joint training} The overall joint training framework can be found in Figure \ref{fig2}. The final loss function is defined as
\begin{equation} \label{eq5}
\begin{aligned}
    \mathcal{L}_{total} = \mathcal{L}_{de} + \mathcal{L}_{ce} + \alpha\mathcal{L}_{cross} + \beta\mathcal{L}_{self}
\end{aligned} 
\end{equation}
Here, $\mathcal{L}_{cross}$ and $\mathcal{L}_{self}$ are the knowledge distillation loss with the cross-encoder and defined as in Eq. \eqref{eq8} and \eqref{eq9} respectively, $\alpha$ and $\beta$ are coefficients for them. Besides, $\mathcal{L}_{de}$ and $\mathcal{L}_{ce}$ are the supervised training loss of the dual-encoder and cross-encoder on the labeled data to maximize the $s(m,e_k)$ for the golden entity $e_k$ in the set of candidates $\lbrace e_1,e_2,...,e_K \rbrace $, the loss can be defined as: 
\begin{equation} \label{eq4}
\begin{aligned}
    & \mathcal{L}_{de} = -s_{de}(m,e_k)+log\sum\limits_{j = 1}^{K}{\rm exp}(s_{de}(m,e_j)) \\
    & \mathcal{L}_{ce} = -s_{ce}(m,e_k)+log\sum\limits_{j = 1}^{K}{\rm exp}(s_{ce}(m,e_j)) \\
\end{aligned}    
\end{equation}

\noindent \textbf{Inference} we only apply the mention-encoder to obtain the mention embeddings, and then retrieve targets directly from pre-computing view embeddings via efficient nearest neighbor search. These view embeddings encompass both global and local views and are generated by the entity-encoder following joint training. Although the size of the entity index may increase due to the number of views, the time complexity can remain sub-linear with the index size due to mature nearest neighbor search techniques \citep{zhang2022multi}. 
\subsection{Hard Negative Sampling} \label{sec3.4}
Hard negatives are effective information carriers for difficult knowledge in distillation. Mainstream techniques for generating hard negatives include utilizing static samples \citep{wu2020scalable} or top-K dynamic samples retrieved from a recent iteration of the retriever \citep{xiong2020approximate,zhan2021optimizing}, but these hard negatives may not be suitable for the current model or are pseudo-negatives~(i.e., unlabeled positives) \citep{qu2021rocketqa}. Aiming to mitigate this issue, we adopt a simple negative sampling method that first retrieves top-N candidates, then randomly samples K negatives from them, which reduces the probability of pseudo-negatives and improves the generalization of the retriever.
\begin{table*} 
\centering
\begin{tabular}{l|cccccccc}
\toprule 
\textbf{Method} & \textbf{R@1} & \textbf{R@2} & \textbf{R@4} & \textbf{R@8} & \textbf{R@16} & \textbf{R@32} & \textbf{R@50} & \textbf{R@64}\\
\midrule
BM25 & - & - & - & - & - & - & - & 69.26 \\
BLINK \citep{wu2020scalable} & - & - & - & - & - & - & - & 82.06 \\
\citet{partalidou2022improving} & - & - & - & - & - & - & 84.28 & - \\
BLINK* & 45.59 & 57.55 & 66.10 & 72.47 & 77.65 & 81.69 & 84.31 & 85.56 \\
SOM \citep{zhang2021understanding}  & - & - & - & - & - & - & - & 89.62 \\
MuVER \citep{ma2021muver} & 43.49 & 58.56 & 68.78 & 75.87 & 81.33 & 85.86 & 88.35 & 89.52 \\
\citet{agarwal2022entity} & 50.31 & 61.04 & 68.34 & 74.26 & 78.40 & 82.02 & - & 85.11 \\
GER \citep{wu2023modeling} & 42.86 & - & 66.48 & 73.00 & 78.11 & 82.15 & 84.41 & 85.65 \\
\midrule
\midrule
MVD (ours) & \textbf{52.51} & \textbf{64.77} & \textbf{73.43} & \textbf{79.74} & \textbf{84.35} & \textbf{88.17} & \textbf{90.43} & \textbf{91.55} \\
\bottomrule 
\end{tabular}
\caption{\label{main-results}
\textbf{Recall@K(R@K)} on test set of ZESHEL, \textbf{R@K} measures the percentage of mentions for which the top-K retrieved entities include the golden entities. The best results are shown in \textbf{bold} and the results unavailable are left blank. * is reproduced by \citet{ma2021muver} that expands context length to 512.
}
\end{table*}

\begin{table*}
\centering
\begin{tabular}{l|ccc|ccc|ccc}
\toprule 
\multicolumn{1}{l|}{\multirow{2}{*}{Method}} & \multicolumn{3}{c|}{AIDA-b} & \multicolumn{3}{c|}{MSNBC} & \multicolumn{3}{c}{WNED-CWEB} \\
& R@10 & R@30 & R@100 & R@10 & R@30 & R@100 & R@10 & R@30 & R@100 \\
\midrule
BLINK & 92.38 & 94.87 & 96.63 & 93.03 & 95.46 & 96.76 & 82.23 & 86.09 & 88.68\\
MuVER & 94.53 & 95.25 & 98.11 & 95.02 & 96.62 & 97.75 & \uline{79.31} & \uline{83.94} & \uline{88.15} \\
MVD (ours)& \textbf{97.05} & \textbf{98.15} & \textbf{98.80} & \textbf{96.74} & \textbf{97.71} & \textbf{98.04} & \textbf{85.01} & \textbf{88.18} & \textbf{91.11} \\
\bottomrule 
\end{tabular}
\caption{\label{wikipedia-results}
\textbf{Recall@K(R@K)} on test set of Wikipedia datasets, best results are shown in \textbf{bold}. Underline notes for the results we reproduce.
}
\end{table*}
\section{Experiments}
\subsection{Datasets}
We evaluate MVD under two distinct types of datasets: three standard EL datasets AIDA-CoNLL \citep{hoffart2011robust}, WNED-CWEB \citep{guo2018robust} and MSNBC \citep{cucerzan2007large}, these datasets are all constructed based on a uniform Wikipedia KB; and a more challenging Wikia-based dataset ZESHEL \citep{logeswaran2019zero}, adopts a unique setup where the train, valid, and test sets correspond to different KBs. Statistics of these datasets are listed in Appendix \ref{a1}.
\subsection{Training Procedure}
The training pipeline of MVD consists of two stages: Warmup training and MVD training. In the Warmup training stage, we separately train dual-encoder and cross-encoder by in-batch negatives and static negatives. Then we initialize the student model and the teacher model with the well-trained dual-encoder and cross-encoder, and perform multi-view enhanced distillation to jointly optimize the two modules following Section \ref{sec3.3}. Implementation details are listed in Appendix \ref{a2}.
\subsection{Main Results}
\noindent \textbf{Compared Methods} We compare MVD with previous state-of-the-art methods. These methods can be divided into several categories according to the representations of entities: BM25 \citep{robertson2009probabilistic} is a sparse retrieval model based on exact term matching. BLINK~\citep{wu2020scalable} adopts a typical dual-encoder architecture that embeds the entity independently into a single fixed-size vector. SOM \citep{zhang2021understanding} represents entities by its tokens and computes relevance scores via the sum-of-max operation~\citep{khattab2020colbert}. Similar to our work, MuVER \citep{ma2021muver} constructs multi-view entity representations to match divergent mentions and achieved the best results, so we select MuVER as the main compared baseline. Besides, ARBORESCENCE \citep{agarwal2022entity} and GER \citep{wu2023modeling} construct mention/entity centralized graphs to learn fine-grained entity representations.

\noindent \textbf{For Zeshel dataset} we compare MVD with all the above models. As shown in Table \ref{main-results}, MVD performs better than all the existing methods. Compared to the previously best performing method MuVER, MVD significantly surpasses in all metrics, particularly in \textbf{R@1}, which indicates the ability to directly obtain the target entity. This demonstrates the effectiveness of MVD, which uses hard negatives as information carriers to explicitly transfer knowledge of multiple fine-grained views from the cross-encoder to better represent entities for matching multiple mentions, resulting in higher-quality candidates for the ranking stage.

\noindent \textbf{For Wikipedia datasets} we compare MVD with BLINK \footnote{BLINK performance is reported in \url{https://github.com/facebookresearch/BLINK}} and MuVER \citep{ma2021muver}. As shown in Table \ref{wikipedia-results}, our MVD framework also outperforms other methods and achieves state-of-the-art performance on AIDA-b, MSNBC, and WNED-CWEB datasets, which verifies the effectiveness of our method again in standard EL datasets.
\subsection{Ablation and Comparative Studies}
\begin{table}
\centering
\begin{tabular}{l|cc}
\toprule 
\textbf{Model} & \textbf{R@1}  & \textbf{R@64}\\
\midrule
MVD & \textbf{51.69}   & \textbf{89.78} \\
\midrule
- w/o multi-view cross-encoder & 50.85   & 89.24 \\
- w/o relevant-view alignment & 51.02   & 89.55 \\
- w/o self-alignment  & 51.21   & 89.43 \\
- w/o cross-alignment  & 50.82  & 88.71 \\
\midrule
- w/o all components & 51.40  & 84.16 \\
\bottomrule 
\end{tabular}
\caption{\label{ablation-results}
Ablation for fine-grained components in MVD on test set of ZESHEL. Results on Wikipedia-based datasets are similar and omitted due to limited space.
}
\end{table}
\begin{table}
\centering
\begin{tabular}{l|cc}
\toprule 
\textbf{Method} & \textbf{R@1} & \textbf{R@64}\\
\midrule
MVD & \textbf{51.69}  & \textbf{89.78} \\
- w/o dynamic distillation & 51.11  & 88.50 \\
- w/o dynamic negatives & 50.26  & 88.46 \\
\midrule
- w/o all strategies & 50.16  &  87.54 \\
\bottomrule 
\end{tabular}
\caption{\label{ablation-joint}Ablation for training strategies in MVD on test set of ZESHEL. 
}
\end{table}
\subsubsection{Ablation Study}
For conducting fair ablation studies and clearly evaluating the contributions of each fine-grained component and training strategy in MVD, we exclude the coarse-grained global-view to evaluate the capability of transferring knowledge of multiple fine-grained views, and utilize Top-K dynamic hard negatives without random sampling to mitigate the effects of randomness on training.

\noindent \textbf{Fine-grained components} ablation results are presented in Table \ref{ablation-results}. When we replace the multi-view representations in the cross-encoder with the original single vector or remove the relevant view selection based on the results of the cross-encoder, the retrieval performance drops, indicating the importance of providing accurate supervised signals for each view of the entity during distillation. Additionally, the removal of cross-alignment and self-alignment results in a  decrease in performance, highlighting the importance of these alignment mechanisms. Finally, when we exclude all fine-grained components in MVD and employ the traditional distillation paradigm based on single-vector entity representation and entity-level soft-labels, there is a significant decrease in performance, which further emphasizes the effectiveness of learning knowledge of multiple fine-grained and mention-relevant views during distillation.
\begin{table}
\centering
\begin{tabular}{lc|cc}
\toprule 
\textbf{Method} & \textbf{View Type} & \textbf{R@1}  & \textbf{R@64}\\
\midrule
BLINK & global & 46.04  & \textbf{87.46} \\
MuVER & global & 36.90  & 80.65 \\
MVD (ours)  & global & \textbf{47.11}  & 87.04 \\
\midrule
BLINK & local & 37.20  & 86.38 \\
MuVER & local & 41.99  &  89.25 \\
MVD (ours) & local & \textbf{51.27}  &  \textbf{90.25} \\
\midrule
MVD (ours)  & global+local & \textbf{52.51}   & \textbf{91.55} \\
\bottomrule 
\end{tabular}
\caption{\label{global-results}Comparison for representing entities from multi-grained views on test set of ZESHEL. Results of BLINK and MuVER are reproduced by us.
}
\end{table}

\noindent \textbf{Training strategies} we further explore the effectiveness of joint training and hard negative sampling in distillation, Table \ref{ablation-joint} shows the results. First, we examine the effect of joint training by freezing the teacher model's parameters to do a static distillation, the retrieval performance drops due to the teacher model's limitation. Similarly, the performance drops a lot when we replace the dynamic hard negatives with static negatives, which demonstrates the importance of dynamic hard negatives for making the learning task more challenging. Furthermore, when both training strategies are excluded and the student model is independently trained using static negatives, a substantial decrease in retrieval performance is observed, which validates the effectiveness of both training strategies in enhancing retrieval performance.
\subsubsection{Comparative Study on Entity Representation}
To demonstrate the capability of representing entities from multi-grained views, we carry out comparative analyses between MVD and BLINK \citep{wu2020scalable}, as well as MuVER \citep{ma2021muver}. These systems are founded on the principles of coarse-grained global-views and fine-grained local-views, respectively.

We evaluate the retrieval performance of both entity representations and present the results in Table \ref{global-results}. The results clearly indicate that MVD surpasses both BLINK and MuVER in terms of entity representation performance, even exceeding BLINK's global-view performance in \textbf{R@1}, despite being a fine-grained training framework. Unsurprisingly, the optimal retrieval performance is attained when MVD employs both entity representations concurrently during the inference process.
\begin{table}[t]
\centering
\begin{tabular}{l|c}
\toprule 
\textbf{Candidate Retriever} & \textbf{U.Acc.}  \\
\midrule
\midrule
\multicolumn{2}{c}{\textit{Base Version Ranker}} \\
\midrule
 BM25 \citep{logeswaran2019zero}& 55.08 \\
 BLINK \citep{wu2020scalable} & 61.34 \\
 SOM \citep{zhang2021understanding} & 65.39 \\
 \citet{agarwal2022entity} & 62.53 \\
 MVD (ours)  & \textbf{66.85} \\
 \midrule
 \midrule
 \multicolumn{2}{c}{\textit{Large Version Ranker}} \\
\midrule
 BLINK \citep{wu2020scalable}& 63.03 \\
 SOM \citep{zhang2021understanding}& 67.14 \\
 MVD (ours) & \textbf{67.84} \\
\bottomrule 
\end{tabular}
\caption{\label{rank-results}
Performance of ranker based on different candidate retrievers on the test set of ZESHEL. \textbf{U.Acc.} means the unnormalized macro accuracy.
}
\end{table}
\section{Further Analysis}
\subsection{Facilitating Ranker's Performance} \label{sec5.1}
To evaluate the impact of the quality of candidate entities on overall performance, we consider two aspects: candidates generated by different retrievers and the number of candidate entities used in inference. First, we separately train BERT-base and BERT-large based cross-encoders to rank the top-64 candidate entities retrieved by MVD. As shown in Table \ref{rank-results}, the ranker based on our framework achieves the best results in the two-stage performance compared to other candidate retrievers, demonstrating its ability to generate high-quality candidate entities for the ranking stage. 

\begin{figure}[t]   
\centering 
\vspace*{-0.6cm}
\includegraphics[width=8cm]{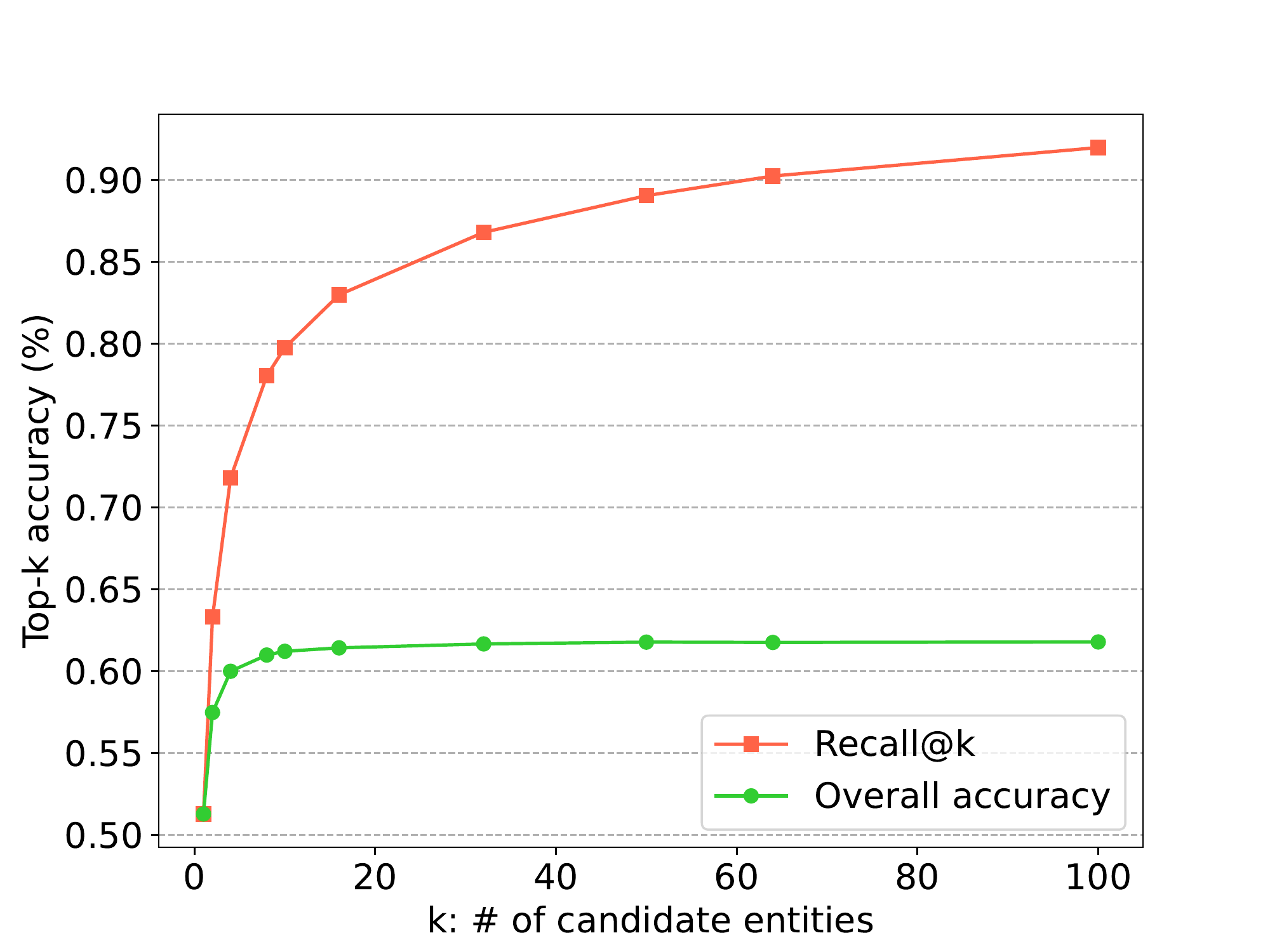}
\caption{Recall and overall micro accuracy based on different number of candidates k.}
\label{fig3}
\end{figure}

Additionally, we study the impact of the number of candidate entities on overall performance, as shown in Figure \ref{fig3}, with the increase of candidates number $k$, the retrieval performance grows steadily while the overall performance is likely to be stagnant. This indicates that it's ideal to choose an appropriate k to balance the efficiency and efficacy, we observe that $k = 16$ is optimal on most of the existing EL benchmarks.
\subsection{Qualitative Analysis}
To better understand the practical implications of fine-grained knowledge transfer and global-view entity representation in MVD, as shown in Table \ref{case-study}, we conduct comparative analysis between our method and MuVER \citep{ma2021muver} using retrieval examples from the test set of ZESHEL for qualitative analysis. 

In the first example, MVD clearly demonstrates its ability to accurately capture the mention-relevant information \textit{Rekelen were members of this movement} and \textit{professor Natima Lang} in the golden entity ``Cardassian dissident movement''. In contrast, MuVER exhibits limited discriminatory ability in distinguishing between the golden entity and the hard negative entity ``Romulan underground movement''. In the second example, Unlike MuVER which solely focuses on local information within the entity, MVD can holistically model multiple mention-relevant parts within the golden entity ``Greater ironguard'' through a global-view entity representation, enabling matching with the corresponding mention ``improved version of lesser ironguard''.
\begin{table*}
\small
\centering
\begin{tabular}{p{4cm}|p{5.3cm}|p{5.7cm}}
\toprule 
\textbf{Mention and Context} & \textbf{Entity retrieved by MVD} & \textbf{Entity retrieved by MuVER}  \\
\midrule
  & Title: Cardassian dissident movement & Title: Romulan underground movement \\
 Rekelen was a member of the \textbf{underground movement} and a student under professor Natima Lang. In 2370, Rekelen was forced to flee Cardassia prime because of her political views.  & The Cardassian dissident movement was a resistance movement formed to resist and oppose the Cardassian Central Command and restore the authority of the Detapa Council. They believed this change was critical for the future of their people. \textbf{Professor Natima Lang, Hogue, and Rekelen were members of this movement in the late 2360s and 2370s}. ...  & The Romulan underground movement was formed sometime prior to the late 24th century on the planet Romulus by a group of Romulan citizens who opposed the Romulan High Command and who supported a Romulan - Vulcan reunification. \textbf{Its methods and principles were similar to those of the Cardassian dissident movement which emerged in the Cardassian Union around the same time}. ...  \\
 \midrule
  & Title: Greater ironguard & Title: Lesser ironguard \\
  Known as the \textbf{improved version of lesser ironguard}, this spell granted the complete immunity from all common, unenchanted metals to the caster or one creature touched by the caster. & Greater ironguard was an arcane abjuration spell that temporarily \textbf{granted one creature immunity from all non-magical metals and some enchanted metals}. \textbf{It was an improved version of ironguard}. The effects of this spell were the same as for "lesser ironguard" except that it also granted immunity and transparency to metals that had been enchanted up to a certain degree. ... & ... \textbf{after an improved version was developed, this spell became known as lesser ironguard}. Upon casting this spell, the caster or one creature touched by the caster became completely immune to common, unenchanted metal. metal weapons would pass through the individual without causing harm. likewise, the target of this spell could pass through metal barriers such as iron bars, grates, or portcullises. ... \\
\bottomrule 
\end{tabular}
\caption{\label{case-study}Examples of entities retrieved by MVD and MuVER, mentions in contexts and mention-relevant information in entities are in \textbf{bold}.}
\end{table*}
\section{Conclusion}
In this paper, we propose a novel Multi-View Enhanced Distillation framework for dense entity retrieval. Our framework enables better representation of entities through multi-grained views, and by using hard negatives as information carriers to effectively transfer knowledge of multiple fine-grained and mention-relevant views from the more powerful cross-encoder to the dual-encoder. We also design cross-alignment and self-alignment mechanisms for this framework to facilitate the fine-grained knowledge distillation from the teacher model to the student model. Our experiments on several entity linking benchmarks show that our approach achieves state-of-the-art entity linking performance. 
\section*{Limitations}
The limitations of our method are as follows:
\begin{itemize}
    \item We find that utilizing multi-view representations in the cross-encoder is an effective method for MVD, however, the ranking performance of the cross-encoder may slightly decrease. Therefore, it is sub-optimal to directly use the cross-encoder model for entity ranking.
    \item Mention detection is the predecessor task of our retrieval model, so our retrieval model will be affected by the error of the mention detection. Therefore, designing a joint model of mention detection and entity retrieval is an improvement direction of our method.
\end{itemize}
\section*{Acknowledgements}
This work is supported by the National Key Research and Development Program of China (NO.2022YFB3102200) and Strategic Priority Research Program of the Chinese Academy of Sciences with No. XDC02030400.
\normalem
\bibliographystyle{acl_natbib}
\bibliography{anthology,acl2023}
\clearpage
\appendix
\section{Appendix}
\label{sec:appendix}
\subsection{Statistics of Datasets} \label{a1}
Table \ref{data-results} shows statistics for ZESHEL dataset, which was constructed based on documents in Wikia from 16 domains, 8 for train, 4 for valid, and 4 for test. 
\begin{table}[htb]
\centering
\begin{tabular}{lcc}
\toprule 
\multicolumn{1}{l}{\multirow{2}{*}{Domain}} & \multicolumn{1}{c}{\multirow{2}{*}{\#Entity}}  & \multicolumn{1}{c}{\multirow{2}{*}{\#Mention}} \\
\\
\hline
 \multicolumn{3}{c}{\textit{Training}} \\
 \hline
 American Football & 31929  & 3898  \\
 Doctor Who & 40281  & 8334  \\
 Fallout & 16992  & 3286  \\
 Final Fantasy & 14044  & 6041  \\
 Military & 104520  & 13063  \\
 Pro Wrestling & 10133  & 1392  \\
 Star Wars & 87056  & 11824  \\
 World of Warcraft & 27677  & 1437  \\
 \textbf{Training} & \textbf{332632} & \textbf{49275} \\
\hline
 \multicolumn{3}{c}{\textit{Validation}} \\
\hline
 Coronation Street & 17809   & 1464 \\
 Muppets & 21344   & 2028 \\
 Ice Hockey & 28684   & 2233 \\
 Elder Scrolls & 21712   & 4275 \\
 \textbf{Validation} & \textbf{89549} & \textbf{10000}  \\
\hline
 \multicolumn{3}{c}{\textit{Testing}} \\
\hline
 Forgotten Realms & 15603   & 1200 \\
 Lego & 10076  &  1199 \\
 Star Trek & 34430  &  4227 \\
 YuGiOh & 10031  &  3374 \\
 \textbf{Testing} & \textbf{70140} & \textbf{10000}  \\
\bottomrule 
\end{tabular}
\caption{\label{data-results}
Statistics of  ZESHEL dataset.
}
\end{table}
Table \ref{four-results} shows statistics for three Wikipedia-based datasets: AIDA, MSNBC, and WNED-CWEB. MSNBC and WNED-CWEB are two out-of-domain test sets, which are evaluated on the model trained on AIDA-train, and we test them on the version of Wikipedia dump provided in KILT \citep{petroni2021kilt}, which contains 5.9M entities.
\begin{table}[htb]
\centering
\begin{tabular}{cccc}
\toprule 
Dataset & \#Mention & \#Entity  \\
\hline
AIDA-train & 18448 & \multirow{5}{*}{5903530}  \\
AIDA-valid & 4791 &  \\
AIDA-test  & 4485 &  \\
\cline{1-2}
MSNBC  & 678 &   \\
WNED-CWEB & 10392 & \\
\bottomrule 
\end{tabular}
\caption{\label{four-results}
Statistics of  three Wikipedia-based datasets.
}
\end{table}

 \subsection{Implementation Details} \label{a2}
For ZESHEL, we use the BERT-base to initialize both the student dual-encoder and the teacher cross-encoder. For Wikipedia-based datasets, we finetune our model based on the model released by BLINK, which is pre-trained on 9M annotated mention-entity pairs with BERT-large. All experiments are performed on 4 A6000 GPUs and the results are the average of 5 runs with different random seeds.

\noindent \textbf{Warmup training} We initially train a dual-encoder using in-batch negatives, followed by training a cross-encoder as the teacher model via the top-k static hard negatives generated by the dual-encoder. Both models utilize multi-view entity representations and are optimized using the loss defined in Eq. \eqref{eq4}, training details are listed in Table \ref{dual-results}.
\begin{table}[ht]
\centering
\begin{tabular}{l|cc}
\toprule 
Hyperparameter & ZESHEL & Wikipedia \\
\midrule
\midrule
\multicolumn{3}{c}{Dual-encoder} \\
\midrule
 Max mention length & 128 & 32 \\
 Max view num & 10 & 5 \\
 Max view length & 40 & 40 \\
 Learning rate & 1e-5 & 1e-5 \\
 Negative num & 63 & 63 \\
 Batch size & 64 &  64 \\
 Training epoch & 40  & 40 \\
 Training time & 4h & 2h \\
\midrule
\midrule
\multicolumn{3}{c}{Cross-encoder} \\
\midrule
 Max input length & 168 & 72 \\
 Learning rate & 2e-5 & 2e-5 \\
 Negative num & 15 & 15 \\
 Batch size & 1 &  1 \\
 Training epoch & 3  & 3 \\
 Training time & 7h & 5h \\
\bottomrule 
\end{tabular}
\caption{\label{dual-results}
Hyperparameters for Warmup training.
}
\end{table}

\noindent \textbf{MVD training} Next, we initialize the student model and the teacher model with the well-trained dual-encoder and cross-encoder obtained from the Warmup training stage. We then employ multi-view enhanced distillation to jointly optimize both modules, as described in Section \ref{sec3.3}. To determine the values of $\alpha$ and $\beta$ in Eq. \eqref{eq5}, we conduct a grid search and find that setting $\alpha = 0.3$ and $\beta = 0.1$ yields the best performance. We further adopt a simple negative sampling method in Sec \ref{sec3.4} that first retrieves top-N candidates and then samples K as negatives. Based on the analysis in Sec \ref{sec5.1} that 16 is the optimal candidate number to cover most hard negatives and balance the efficiency, we set it as the value of K; then to ensure high recall rates and sampling high quality negatives, we search from a candidate list [50, 100, 150, 200, 300] and eventually determine N=100 is the most suitable value. The training details are listed in Table \ref{joint-results}.

\begin{table}[ht]
\centering
\begin{tabular}{l|cc}
\toprule 
Hyperparameter & ZESHEL & Wikipedia \\
\midrule
 Max mention length & 128 & 32 \\
 Max view num & 10 & 5 \\
 Max view length & 40 & 40 \\
 Max cross length & 168 & 72 \\
 Learning rate & 2e-5 & 2e-5 \\
 Negative num & 15 & 15 \\
 Batch size & 1 &  1 \\
 Training epoch & 5  & 5 \\
 Training time & 15h & 6h \\
\bottomrule 
\end{tabular}
\caption{\label{joint-results}
Hyperparameters for MVD training.
}
\end{table}
\noindent \textbf{Inference} MVD employs both local-view and global-view entity representations concurrently during the inference process, details are listed in Table \ref{inference-results}.
\begin{table}[ht]
\centering
\begin{tabular}{l|cc}
\toprule 
Hyperparameter & ZESHEL & Wikipedia \\
\midrule
Local-view length & 40 & 40 \\
Global-view length & 512 & 128 \\
 Avg view num & 16 & 6 \\
\bottomrule 
\end{tabular}
\caption{\label{inference-results}
Hyperparameters for Inference.
}
\end{table}
\end{document}